\documentclass[submission,copyright,creativecommons]{eptcs}
\providecommand{\event}{ICLP 2023} 

\usepackage{iftex}
\usepackage{amsmath}
\usepackage{graphicx}
\usepackage{multirow}
\usepackage{todonotes}
\usepackage{url}
\usepackage{xspace}
\usepackage[nonumberlist, acronym,]{glossaries}
\usepackage{amsfonts}
\usepackage{booktabs}
\usepackage{listings}
\usepackage{cleveref}
\crefname{enumi}{}{}
\crefname{equation}{}{}
\newcommand{\CalP}{\mathcal{P}}
\newcommand{\CalC}{\mathcal{C}}
\newcommand{\CalA}{\mathcal{A}}

\newcommand{\CalT}{\mathcal{T}}

\newcommand{\ground}{\text{g}}
\newcommand{\defneg}{\mathit{not}\;}

\newcommand{\Ploc}{\CalP_\loc}
\newcommand{\loc}{\textsf{loc}\xspace}
\newcommand{\ham}{\text{hamburg}\xspace}
\newcommand{\hamharbor}{\text{hamburgHarbor}\xspace}
\newcommand{\prodLoc}{\textsf{productionLoc}\xspace}
\newcommand{\wareLoc}{\textsf{warehouseLoc}\xspace}
\newcommand{\prodPlan}{\textsf{productionPlan}\xspace}
\newcommand{\rootPart}{\textsf{root}\xspace}
\newcommand{\super}{\textsf{Super}\xspace}

\newcommand{\Ac}{\textsf{aCountry}\xspace}
\newcommand{\Bc}{\textsf{bCountry}\xspace}
\newcommand{\Cc}{\textsf{cCountry}\xspace}

\newcommand{\aPa}{\textsf{aP}\xspace}
\newcommand{\bPa}{\textsf{bP}\xspace}
\newcommand{\cPa}{\textsf{cP}\xspace}

\newcommand{\aHa}{\textsf{aH}\xspace}
\newcommand{\bHa}{\textsf{bH}\xspace}

\newcommand{\country}{\textsf{country}\xspace}
\newcommand{\location}{\textsf{location}\xspace}
\newcommand{\productionLocation}{\prodLoc} 
\newcommand{\locatedIn}{\textsf{locatedIn}\xspace}
\newcommand{\invLocatedIn}{\textsf{invalidLocatedIn}\xspace}
\newcommand{\invalidLocatedInTwoCountries}{\textsf{invalidLocatedInTwoCountries}\xspace}

\newcommand{\transportMean}{\textsf{transportMean}\xspace}
\newcommand{\plane}{\textsf{plane}\xspace}
\newcommand{\ship}{\textsf{ship}\xspace}
\newcommand{\truck}{\textsf{truck}\xspace}
\newcommand{\transportMeanAtSite}{\textsf{transportMeanAtSite}\xspace}

\newcommand{\canTransport}{\textsf{canTransport}\xspace}
\newcommand{\partProduceableAt}{\textsf{partProduceableAt}\xspace}
\newcommand{\productionPart}{\textsf{part}\xspace}
\newcommand{\variablePart}{\textsf{Part}\xspace}
\newcommand{\partAbbrv}{\textsf{p}}
\newcommand{\transportRoute}{\textsf{transportRoute}\xspace}
\newcommand{\productionPlan}{\textsf{productionPlan}\xspace}
\newcommand{\canBeTransportedFromTo}{\textsf{canBeTransportedFromTo}\xspace}
\newcommand{ \intrasiteTransport}{\textsf{intrasiteTransport}\xspace}
\newcommand{\PL}{\textsf{PL}}
\newcommand{\partProducedAtLocation}{\textsf{producedAtLoc}\xspace}

\newcommand{\via}{\textsf{via}}
\newcommand{\variableVia}{\textsf{Via}}
\newcommand{\direct}{\textsf{direct}\xspace}
\newcommand{ \From}{\textsf{From}\xspace}
\newcommand{ \To}{\textsf{To}\xspace}
\newcommand{ \TM}{\textsf{TM}\xspace}
\newcommand{ \Distance}{\textsf{D}\xspace}

\newcommand{\variableLocation}{\textsf{Location}}
\newcommand{\variableCountry}{\textsf{Country}}

\newcommand{\partProducedAt}{\textsf{partProducedAt}}
\newcommand{\transportPath}{\textsf{path}}
\newcommand{\minimize}{\textsf{minimize}}

\newcommand{\Xvar}{\textsf{X}}
\newcommand{\Yvar}{\textsf{Y}}

\glsdisablehyper
\newacronym{ADTS}{ADTS}{Airbus Digital Trust Solutions}
\newacronym{AI}{AI}{Artificial Intelligence}
\newacronym{API}{API}{Application Programming Interface}
\newacronym{ARM}{ARM}{Advanced RISC Machine}
\newacronym{ASP}{ASP}{Answer Set Programming}
\newacronym{ASPIRE}{ASPIRE}{Application and Software Product IM Repositories}
\newacronym{AWS}{AWS}{Amazon Web Services}
\newacronym{BDM}{BDM}{Behavioural Domain Model}
\newacronym{CA}{CA}{Certificate Authority}
\newacronym{CAD}{CAD}{Computer Aided Design}
\newacronym{CAE}{CAE}{Computer Aided Engineering}
\newacronym{CDO}{CDO}{Connected Data Objects}
\newacronym{CI}{CI}{Continuous Integration}
\newacronym{CnC}{C\&C}{Command and Control}
\newacronym{COTS}{COTS}{Commercial off-the-shelf}
\newacronym{CPU}{CPU}{Central Processing Unit}
\newacronym{CRT}{CRT}{Central Research and Technology}
\newacronym{DB}{DB}{Database}
\newacronym{DDMS}{DDMS}{Digital Design, Manufacturing and Services}
\newacronym{DSL}{DSL}{Domain-Specific Language}
\newacronym{CSS}{CSS}{Cascading Style Sheets}
\newacronym{DoE}{DoE}{Design of Experiments}
\newacronym{EMF}{EMF}{Eclipse Modelling Framework}
\newacronym{EMT}{EMT}{Eclipse Modelling Tool}
\newacronym{epl}{EPL}{Eclipse Public License}
\newacronym{ERP}{ERP}{Enterprise Resource Planning}
\newacronym{ESXi}{ESXi}{Elastic Sky X integrated}
\newacronym{EWIS}{EWIS}{Electrical Wiring and Interconnection System}
\newacronym{FaaS}{FaaS}{Function as a Service}
\newacronym{FEM}{FEM}{Finite Element Method}
\newacronym{FMU}{FMU}{Functional Mock-up Unit}
\newacronym{FMI}{FMI}{Functional Mock-up Interface}
\newacronym{FPGA}{FPGA}{Field Programmable Gate Arrays}
\newacronym{FIFO}{FIFO}{First-in, First-out}
\newacronym{gmf}{GMF}{Graphical Modeling Framework}
\newacronym{GPU}{GPU}{Graphics Processing Unit}
\newacronym{GUI}{GUI}{Graphical User Interface}
\newacronym{HTML}{HTML}{Hypertext Markup Language}
\newacronym{HTTP}{HTTP}{Hypertext Transfer Protocol}
\newacronym{HTTPS}{HTTPS}{Hypertext Transfer Protocol Secure}
\newacronym{IaaS}{IaaS}{Infrastructure as a Service}
\newacronym{ICT}{ICT}{Information \& Communications Technology}
\newacronym{ide}{IDE}{Integrated Development Environment}
\newacronym{IF}{IF}{Interface}
\newacronym{IIS}{IIS}{Internet Information Services}
\newacronym{IM}{IM}{Information Management}
\newacronym{INCOSE}{INCOSE}{International Council on Systems Engineering}
\newacronym{IoT}{IoT}{Internet of Things}
\newacronym{IP}{IP}{Intellectual Property}
\newacronym{IT}{IT}{Information Technology}
\newacronym{JSON}{JSON}{JavaScript Object Notation}
\newacronym{KPI}{KPI}{Key Performance Indicator}
\newacronym{LDAP}{LDAP}{Lightweight Directory Access Protocol}
\newacronym{ManagHAS}{ManagHAS}{Manage Hardware Applications Servers}
\newacronym{MBSE}{MBSE}{Model-based Systems Engineering}
\newacronym{MSA}{MSA}{Micro Service Architecture}
\newacronym{NRC}{NRC}{Non-Recurring Cost}
\newacronym{OCL}{OCL}{Object Constraint Language}
\newacronym{ODTS}{ODTS}{Obeo Designer Team Server}
\newacronym{OoC}{OoC}{Out-of-Cycle}
\newacronym{osgi}{OSGi}{Open Services Gateway initiative}
\newacronym{PaaS}{PaaS}{Platform as a Service}
\newacronym{PDM}{PDM}{Product Data Management}
\newacronym{PKI}{PKI}{Public Key Infrastricture}
\newacronym{PLM}{PLM}{Product Lifecycle Management}
\newacronym{pom}{POM}{Project Object Model}
\newacronym{RAM}{RAM}{Random-Access Memory}
\newacronym{RC}{RC}{Recurring Cost}
\newacronym{RDF}{RDF}{Resource Description Framework}
\newacronym{REST}{REST}{Representational State Transfer}
\newacronym{rcp}{RCP}{Rich Client Platform}
\newacronym{S2I}{S2I}{Source-2-Image}
\newacronym{SC}{SC}{Smart Component}
\newacronym{SCM}{SCM}{Smart Component Model}
\newacronym{SHACL}{SHACL}{Shapes Constraint Language}
\newacronym{SLA}{SLA}{Service Level Agreement}
\newacronym{SOA}{SOA}{Service-oriented Architecture}
\newacronym{SSO}{SSO}{Single Sign-On}
\newacronym{SVG}{SVG}{Scalable Vector Graphics}
\newacronym{SVO}{SVO}{Subject Verb Object}
\newacronym{TBD}{TBD}{To Be Defined}
\newacronym{UI}{UI}{User Interface}
\newacronym{URL}{URL}{Uniform Resource Locator}
\newacronym{UUID}{UUID}{Universally Unique Identifier}
\newacronym{VHTNG}{VHTNG}{Virtual Hybrid Testing Next Generation}
\newacronym{VPM}{VPM}{Virtual Product Model}
\newacronym{W3C}{W3C}{World Wide Web Consortium}
\newacronym{xmi}{XMI}{XML Metadata Interchange}
\newacronym{XML}{XML}{Extensible Markup Language}
\newacronym{YAML}{YAML}{\textbf{Y}AML \textbf{A}in't \textbf{M}arkup \textbf{L}anguage}
\newacronym{SysML}{SysML}{Systems Modeling Language}
\newacronym{RnT}{R\&T}{Research \& Technology}
\newacronym{MnT}{M\&T}{Methods \& Tools}

\newacronym{BDD}{BDD}{Block Definition Diagram}
\newacronym{BFO}{BFO}{Basic Formal Ontolology}
\newacronym{BPMN}{BPMN}{Business Process Model and Notation}
\newacronym{CWA}{CWA}{Closed World Assumption}
\newacronym{DOE}{DOE}{Design of Experiment}
\newacronym{DRIFT}{DRIFT}{Do It Right the First Time}
\newacronym{DT}{DT}{Digital Twin}
\newacronym{GOPPRR}{GOPPRR}{graph, object, port, property, relationship, role}
\newacronym{IRI}{IRI}{Internationalized Resource Identifier}
\newacronym{KerML}{KerML}{Kernel Modeling Language}
\newacronym{LSP}{LSP}{Language Server Protocol}
\newacronym{MMS}{MMS}{Model Management Server}
\newacronym{MOF}{MOF}{Meta Object Facility}
\newacronym{MSN}{MSN}{Manufacturer Serial Number}
\newacronym{OMG}{OMG}{Object Management Group}
\newacronym{OML}{OML}{Ontological Modeling Language}
\newacronym{OpenMBEE}{OpenMBEE}{Open Model Based Engineering Environment}
\newacronym{OSLC}{OSLC}{Open Services for Lifecycle Collaboration}
\newacronym{OSMC}{OSMC}{Open Source Modelica Consortium}
\newacronym{OWA}{OWA}{Open World Assumption}
\newacronym{OWL}{OWL}{Web Ontology Language}
\newacronym{PAM}{PAM}{Parametric Analysis Definition Model}
\newacronym{PDP}{PDP}{Product Development Plan}
\newacronym{QVT}{QVT}{Query/View/Transformation}
\newacronym{RFLP}{RFLP}{Requirements-Functional-Logical-Physical}
\newacronym{SBRV}{SBRV}{Semantics Of Business Vocabulary And Business Rules}
\newacronym{SECAM}{SECAM}{Systems Engineering Core Architecture Model}
\newacronym{SDLC}{SDLC}{System Development Life Cycle}
\newacronym{SPARQL}{SPARQL}{SPARQL Protocol and RDF Query Language}
\newacronym{SQL}{SQL}{Structured Query Language}
\newacronym{SPO}{SPO}{Subject Predicate Object}
\newacronym{SWRL}{SWRL}{Semantic Web Rule Language}
\newacronym{TWC}{TWC}{TeamWork Cloud}
\newacronym{UML}{UML}{Unified Modeling Language}
\newacronym{VA}{VA}{Variant Analysis}

\newacronym{iso}{ISO}{International Organization for Standardization}
\newacronym{csi}{CSI}{Common Simulation Infrastructure}
\newacronym{mossec}{MoSSEC}{Modelling and Simulation information in a collaborative Systems Engineering Context}
\newacronym{amo}{AMO}{Airbus MOdels Management Team}

\ifpdf
  \usepackage{underscore}         
  \usepackage[T1]{fontenc}        
\else
  \usepackage{breakurl}           
\fi

\newcommand{\astfootnote}[1]{%
\let\oldthefootnote=\thefootnote%
\setcounter{footnote}{0}%
\renewcommand{\thefootnote}{\fnsymbol{footnote}}%
\footnote{#1}%
\let\thefootnote=\oldthefootnote%
}

\title{A Logic Programming Approach to Global Logistics in a Co-Design Environment}
\author{Emmanuelle Dietz\textsuperscript{1}, Tobias
  Philipp\textsuperscript{2}, Gerrit Schramm\textsuperscript{1}, Andreas Zindel\textsuperscript{1}\astfootnote{The authors are in alphabetical order.}
\institute{\textsuperscript{1} Airbus Central Research \& Technology, Germany \\ \textsuperscript{2} secunet Security Networks AG, Germany}
\email{\quad {\{firstname.lastname\}@{airbus.com,secunet.com}}}
}

\newcommand{\titlerunning}{A Logic Programming Approach to Global Logistics in a Co-Design Environment}
\newcommand{\authorrunning}{E. Dietz, T. Philipp, G. Schramm, A. Zindel}

\hypersetup{
  bookmarksnumbered,
  pdftitle    = {\titlerunning},
  pdfauthor   = {\authorrunning},
  pdfsubject  = {EPTCS},               
  pdfkeywords = {Logic Programming, Co-Design, Global Logistics, Knowledge Graphs, Ontologies} 
}

\begin{document}
\maketitle

\begin{abstract}
In a co-design environment changes need to be integrated quickly and in an automated manner.
This paper considers the challenge of creating and optimizing a global logistics system for the construction of a passenger aircraft within a co-design approach with respect to key performance indicators (like cost, time or resilience).
The product in question is an aircraft, comprised of multiple components, manufactured at multiple sites worldwide.
The goal is to find an optimal way to build the aircraft taking into consideration the requirements for its industrial system.
The main motivation for approaching this challenge is to develop the industrial system in tandem with the product and making it more resilient against unforeseen events, reducing the risks of bottlenecks in the supply chain.
This risk reduction ensures continued efficiency and operational success.
To address this challenging and complex task we have chosen \gls{ASP} as the modeling language, formalizing the relevant requirements of the investigated industrial system. 
The approach presented in this paper covers three main aspects: the extraction of the relevant information from a knowledge graph, the translation into logic programs and the computation of existing configurations guided by optimization criteria.
Finally we visualize the results for an effortless evaluation of these models.
Internal results seem promising and yielded several new research questions for future improvements of the discussed use case.
\end{abstract}

\section{Introduction}\label{sect:intro}
Co-design is a holistic approach (\cite{mitchell2015codesign}) for the development of new products in tandem with an existing industrial system, and is essential for the development of innovative and complex products.
Consider the scenario that a product development department chooses a material that cannot be processed by the existing industrial system.
This may lead to significant cost increases and bottlenecks later in the lifecycle.
In contrast, a co-design enabled environment encourages trade-offs and suggests to adapt either the product or the industrial system.
The results of that trade-off are integrated into the future developments, before the product is ready-to-market.

The first step towards co-design is to determine if a product can be produced by the existing industrial system.
This assessment is especially difficult for products --- like aircrafts --- where not only the product itself is complex but which is also manufactured by a global industrial system in an intricate logistics network.
Therefore, we propose the following approach: a data model to represent the industrial system and the product, means to capture the data about these systems, a method to create valid system architectures, ways to calculate key performance indicators and lastly visualize the results.

Creating a unified data model is a challenging  task since the co-design approach requires to combine different data domains together.
We propose an \gls{RDF} based ontology to create the data model.
This ontology needs to be instantiated with the available company knowledge captured in multiple sources, resulting in a co-design knowledge graph.
Since the data can change anytime the attached domains develop their respective systems further the knowledge graph needs to be as lean as possible to quickly adapt to the evolving environment.
After each iteration of the 'to be developed' product new variants of the industrial system need to be created and evaluated. 
A naive approach would expand all possible variants and evaluate each of them.
However, given a sufficiently complex system, this would soon extend the capabilities of even the most advanced computer systems.\footnote{A product with $70$ parts, $16$ possible manufacturers and $2$ manufactures per part yields $10^{166}$ variants.}

We address this challenging task by applying \gls{ASP}, an expressive, logic-based declarative modeling language and problem-solving framework for hard computational problems~(\cite{Niemel1999LogicPW,GebserKaminskiKaufmannSchaub12}).
Given that these techniques require deep knowledge about the deployed technologies this process needs to be as seamless and automated as possible to be suitable for a co-design framework.
Thus, it is supported by a \gls{CI} based process to integrate company knowledge into \gls{ASP} statements automatically. 
For a straightforward evaluation the results are visualized.

The paper is structured as follows: After the preliminaries,
Section~\ref{sect:loginaction} illustrates the structure of the data as a knowledge graph, the extraction of the relevant information and its conversion to logic program facts. 
Next, the (choice) rules and integrity constraints for finding possible logistics configurations optimized with respect to certain performance indicators and the visual representation of these configurations are presented.
Section~\ref{sect:lessonslearned} summarizes some lessons learned.

\section{Background}
The proposed framework is based on a variety of modules that exploit the technological advancements in Software Engineering and Knowledge Representation and Reasoning.
This section introduces the preliminaries for knowledge extraction and computation.

\subsection{Semantic Web Technologies}
\label{sec:semanticIntegration}

The \gls{W3C} defined a standardized technology stack for representing data in a semantic way.
These technologies can be used to create ontologies that model the investigated system and populating them with entities to build a knowledge graph.

In our context an ontology is defined as \textit{a formal, explicit specification of shared conceptualization} (\cite{gruber1993portable}).
\textit{Formal} because an ontology needs to be machine-readable, \textit{explicit specification} demands the capability of defining concepts, properties, constraints and axioms, \textit{shared} as it represents consensual knowledge while \textit{conceptualization} defines an ontology as an abstract model, a simplified view of the system.


The \gls{W3C} technology stack proposes \gls{RDF} as the standard model to represent information in the form of \gls{SPO} triples (\cite{w3c2014rdf}).
To be valid \gls{RDF} each part of the triple has to be an \gls{IRI}, a data type literal --- that stores discrete values --- or a blank node.
Blank nodes are either subjects or objects that do not have an \gls{IRI} or literal.
\gls{RDF} can be serialized into a variety of formats such as \textit{Turtle}, \textit{RDF/XML} or \textit{JSON-LD} depending on the consuming interface (\cite{w3c2014turtle}).
              
To foster re-usability several standardized \gls{RDF} vocabularies are available.
The \glsdesc{OWL} is an \gls{RDF} vocabulary that allows the creation of ontologies (\cite{w3c2012owl}).
\gls{OWL} implements classes (\texttt{owl:Class}), properties (e.g. \texttt{owl:ObjectProperty}, , \texttt{owl:Datatype-}\\
\texttt{Property}), constraints (e.g. \texttt{owl:minCardinality}) and axioms (e.g. \texttt{owl:equivalentClass}).
An ontology forms one part of a knowledge graph, called the terminological component (\emph{TBox}). 
The other part, that contains instances of the classes defined in the \emph{TBox}, is called assertion component (\emph{ABox}) (\cite{gruber1993portable}).
Tools like \textit{Protégé} support the creation of \gls{OWL} ontologies by providing a graphical editor to implement concepts and relations (\cite{stanford2022protege}).

\gls{OWL} can be split into several fragments with varying levels of expressiveness.
These fragments, like \gls{OWL}-DL, provide a subset of constructs of the full implementation of \gls{OWL}.
Since \gls{OWL}-DL implements the $\mathcal{SROIQ(D)}$ Description Logic (hence the suffix DL) (\cite{horrocks2006sroiq}) reasoning engines can ingest knowledge graphs backed by an \gls{OWL}-DL ontology in order to infer new knowledge based on existing facts. 

\gls{SWRL} provides a way to express these inferences rules.
These rule sets can be used by the reasoning engines to create new knowledge (\cite{w3c2004swrl}).
\gls{OWL}-DL (like all \gls{OWL} fragments) is using the \gls{OWA} for reasoning.
Possibly, reasoners might not terminate, since \gls{OWL} and its fragments are undecidable.


In order to access data stored in the knowledge graphs specialized data stores and query languages have been developed.
\gls{SPARQL} is the \gls{W3C} recommendation for querying knowledge graphs implemented with \gls{RDF} (\cite{w3c2013sparql}).
\gls{SPARQL} provides expressions and features of traditional query languages for relation databases like the \gls{SQL}.
However, since \gls{SPARQL} is implemented as a graph pattern matching language, it provides additional methods for traversing \gls{RDF} graphs, querying multiple graphs at once and constructing sub-graphs.
\gls{SPARQL} is running under the \gls{CWA}, either it matches a graph pattern or not.
Executing a query will always yield a result, even if empty.
An extension of \gls{SPARQL} called \gls{SPARQL} Update allows to manipulate the data in the knowledge graph by inserting, updating and deleting triples with the help of a query (\cite{w3c2013sparqlupdate}).

Specialized data stores for \gls{RDF} data are called triple stores. 
These stores come in two varieties: they are either storing \gls{RDF} data natively, or convert them into their own, internal data format.
Usually these storage solutions come with tools, \glspl{API} and \gls{SPARQL} endpoints to extract and manipulate \gls{RDF} triples.
One such storage solution is MarkLogic (\cite{marklogic2019semantics}).
MarkLogic is a NoSQL document store that stores all data in \gls{XML} documents. 
In order to store \gls{RDF} data the \gls{RDF} triples are put into documents.
Internally the provided \gls{SPARQL} endpoints access the data stored in the documents.

MarkLogic provides reasoning at query run time; so called backward-chaining inference (\cite{marklogic2019semantics}).
However, this is resource intensive. 
To facilitate fast query execution times, reasoning via ontologies or \gls{SWRL} rules should be done before executing the query, explicitly materializing the inferred facts.
Not only due to the additional run time of the reasoning step but also because of the undecidability of \gls{OWL}.

We focus on two areas of Semantic Web technologies:
unify data from several diverse sources supported by an ontology and generate new knowledge with the help of reasoning to keep the resulting knowledge graph as concise and easy to maintain as possible.

\subsection{Logic Programming and Stable Model Semantics}

We introduce the general notation and terminology used in this paper and assume the reader to be familiar with \gls{ASP}~(\cite{Niemel1999LogicPW,GebserKaminskiKaufmannSchaub12}) and the stable model semantics~(\cite{GelfondLifschitz91}). The interested reader is referred to~\cite{BrewkaEiterTruszczynski11} and \cite{JanhunenNimelae2016}.


We consider the countable set of \textit{terms} $\CalT = \{t_1, \dots, t_n\}$ that consists only of constants and variables with a total order $\leq$ over the elements in $\CalT$.  
An \textit{atom} is an expression $p(t_1,\dots t_m)$ where  $p$ is a predicate,  $m \geq 0$ and $t_1,\dots t_m \subseteq \CalT$. 
$\CalA$ is a fixed, finite and non-empty set of \textit{atoms}.
A \textit{ground atom} is an atom with only constants.
If $A$ is an atom, then $A$ and $\defneg A$ are \textit{literals}, the \textit{positive literal} and the \textit{negative literal}, respectively. 
A \textit{ground formula} is a formula with only constants.
Consider the example where $\prodLoc$ and $\wareLoc$ are predicates, $\ham$ and $\hamharbor$ are constants and $X$ is a variable: 
$\prodLoc(\ham)$ and $\wareLoc(\hamharbor)$ are ground atoms, $\defneg\prodLoc(\hamharbor)$ is a negative ground literal and $\prodLoc(X)$ is a non-ground atom.
A (normal) \textit{rule} $r$ is of the form
\begin{equation}\small
A \leftarrow  A_1, \dots, A_m, \defneg A_{m+1}, \dots, \defneg A_n. \label{rule}  
\end{equation}
where 
$A$ and $A_i$ with $1 \leq i \leq n$ are atoms. 
The \textit{head} of $r$ is denoted as $H(r)=A$.
The subformula to the right of the implication symbol is called \textit{body} of $r$ and is denoted as $B(r)=A_1, \dots, A_m, \defneg A_{m+1}, \dots, \defneg A_n$.
The set of all positive literals in $B(r)$ is denoted as $B^+(r)= \{A_1, \dots, A_m\}$ and the set of all negative literals in $B(r)$ is denoted as $B^-(r)=\{\defneg A_{m+1}, \dots, \defneg A_n\}$.
A rule is \textit{safe} if each variable in~$r$ occurs in $B^+(r)$. 
A rule is \textit{ground} if no variable occurs in $r$.
A \textit{fact} is a ground rule with empty body, i.e.\ $n=0$.
The rule $\loc(X) \leftarrow \prodLoc(X)$ is safe, but neither ground nor a fact. 

A \textit{logic program} $\CalP$ is a (finite) set of (normal) {rules}.
For any program $\CalP$, let $\CalC$ be the set of all constants appearing in $\CalP$.
In the sequel, we assume for all $\CalP$ it holds that $\CalC_{\CalP} \not= \emptyset$.
The ground program $\ground\CalP$ is the set of rules $r\sigma$ obtained by applying to each rule $r \in \CalP$, all possible substitutions $\sigma$ from the variables in $r$ to the elements of $\CalC_\CalP$.

Program $\Ploc$ consists of the following rules, where $\CalC_{\Ploc} = \{\ham,\hamharbor\}$:
\[\small
\begin{array}{@{\hspace{0cm}}l}
\loc(X) \leftarrow \prodLoc(X). \quad \prodLoc(\ham). \quad \wareLoc(\hamharbor).
\end{array}
\]
The ground program $\ground\Ploc$ consists of the following rules:
\[
\small
\begin{array}{@{\hspace{0cm}}rl@{\hspace{0.1cm}}l}
\loc(\ham) & \leftarrow \prodLoc(\ham). & \quad\prodLoc(\ham). \\
\loc(\hamharbor) & \leftarrow \prodLoc(\hamharbor). & \quad \wareLoc(\hamharbor).
\end{array}
\]
The set of all atoms in $\ground\CalP$ is denoted by $\CalA_\CalP$.
An \textit{interpretation} $I$ of a program $\CalP$ is a mapping of $\CalA_\CalP$ to the set of truth values $\{\top,\bot\}$, where $\top$ means \textit{true} and $\bot$ means \textit{false}. 
Given a ground rule r, $I(r)=\top$ denotes that the interpretation $I$ maps $r$ to $\top$ according to the corresponding logic. 
An interpretation $I$ is a \textit{model} of $\CalP$ where for each rule $r$ occurring in $\ground\CalP$ it holds that $I(r)=\top$.
Consider $\ground\Ploc$ again. 
\[
\small
\begin{array}{ll@{\hspace{0.1cm}}l}
\CalA_{\Ploc} & = \{ & \loc(\ham), \  \prodLoc(\ham),\
\loc(\hamharbor),\\
&&  \prodLoc(\hamharbor),\ \wareLoc(\hamharbor) \ \}.
\end{array}
\]
$I_1 = \CalA_{\Ploc}$, $I_2 = \{\loc(\ham), \prodLoc(\ham),\wareLoc(\hamharbor) \}$ and $I_3 = \emptyset$ are interpretations of $\Ploc$, but  only $I_1$ and $I_2$ are models of $\Ploc$.
An interpretation $I \subseteq \CalA_\CalP$ \textit{satisfies} a ground rule $r$ iff $H(r) \cap I \not = \emptyset$ whenever $B^+(r) \subseteq I$ and $B^- \cap I = \emptyset$. $I$ satisfies a ground program $\CalP$, if each $r \in \CalP$ is satisfied by $I$. A non-ground rule $r$ (resp. a program $\CalP$) is satisfied by an interpretation $I$ if $I$ satisfies all ground instances of $r$ (resp. $\ground\CalP$).
$I$ is an \textit{answer set} (also called \textit{stable model}) of $\CalP$ iff $I$ is the subset-minimal set satisfying the \textit{Gelfond-Lifschitz reduct}: $\CalP^I = \{ H(r) \leftarrow B^+(r) \mid I \cap B^-(r) = \emptyset, r \in \ground\CalP \}$.
$I_2$ is the only answer set of $\Ploc$.

The following two syntactic extensions are commonly used in \gls{ASP}.
\begin{align}
&\leftarrow  A_1, \dots, A_m, \defneg A_{m+1}, \dots, \defneg A_n. \label{ic}\\
&\min\{ A :  A_1, \dots, A_m, \defneg A_{m+1}, \dots, \defneg A_n\}\max. \label{choice}
\end{align}
(\ref{ic}) and (\ref{choice}) are called \textit{integrity constraint} and \textit{choice rule}, respectively. 
Similar as above, $A$ and $A_i$ with $1 \leq i \leq n$ are atoms.
Intuitively, an integrity constraint represents an undesirable situation, i.e. $ A_1, \dots, A_m, \defneg A_{m+1}, \dots, \defneg A_n$ should be evaluated to \textit{false}. 
\textit{No location can be both a warehouse and a production location} can be expressed as $\ \leftarrow \prodLoc(X), \wareLoc(X)$.
In choice rules, $\min$ and $\max$ are non-negative expressions, and $\{ A :  A_1, \dots, A_m, \defneg A_{m+1}, \dots, \defneg A_n\}$ denotes the set of all ground instantiations of $A$,  governed through $\{ A :  A_1, \dots, A_m, \defneg A_{m+1}, \dots, \defneg A_n\}$. 
Intuitively, an interpretation satisfies a choice rule if $\min \leq N \leq \max$ holds, where $N$ is the cardinality of any subset of $\{ A :  A_1, \dots, A_m, \defneg A_{m+1}, \dots, \defneg A_n\}$.
\textit{At least one location to be a production location} can be expressed as $1\{\prodLoc(X): \loc(X)\}$. 

\begin{figure}
	\includegraphics[width=\textwidth]{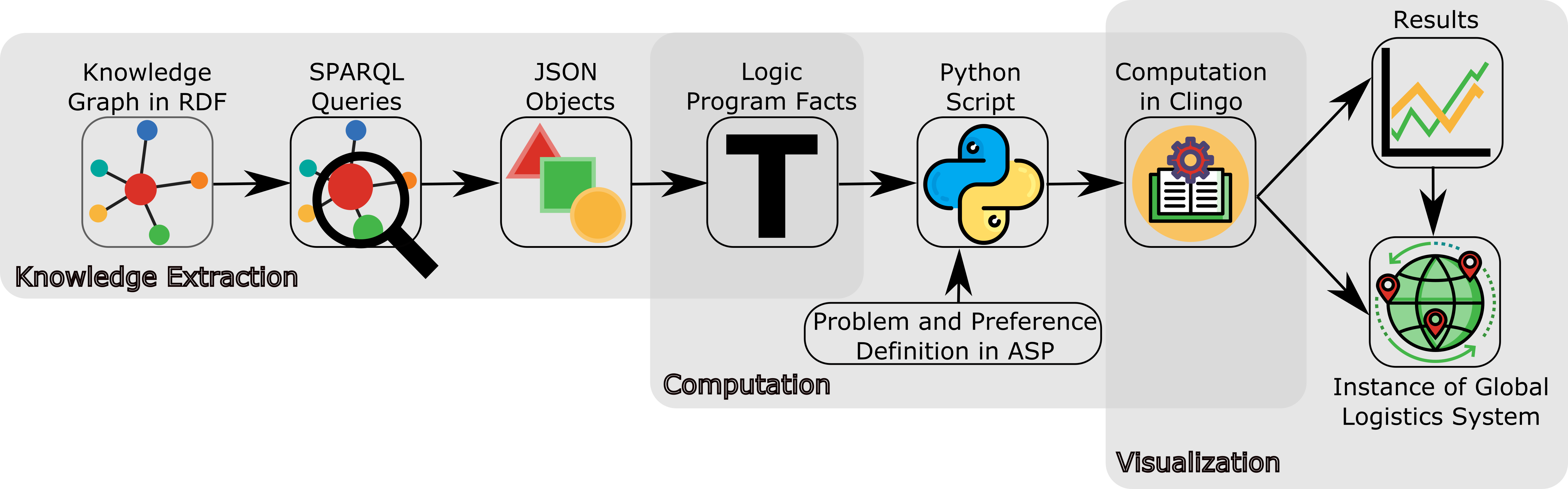}
\caption{Knowledge extraction, computation and visualization for a global logistics framework.
\label{ref:overview-process}}
\end{figure}

\section{Logistics in Action}\label{sect:loginaction}
 \begin{figure}[t]
	\centering
	\includegraphics[width=\textwidth]{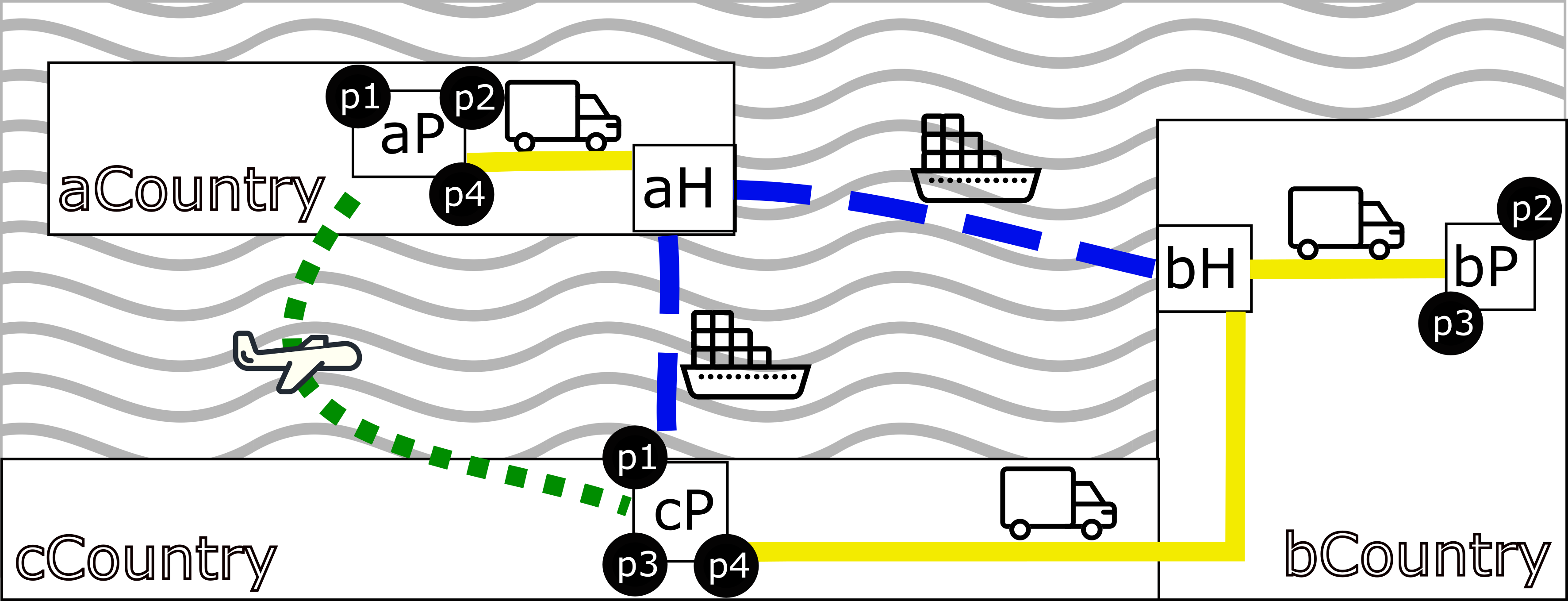}
	\caption{A logistics system where the gray waves, the black rectangles and the black squares represent a sea, countries and locations, respectively.  The lines represent transport routes.
	\label{fig:logistics-example}}
\end{figure}

Figure~\ref{ref:overview-process} shows the proposed framework:
The first box on the left contains the information about the global logistics system as a knowledge graph in \gls{RDF} on a triple store. 
Predefined queries specified in \gls{SPARQL} allow the extraction of relevant data resulting in \gls{JSON} objects that are transposed with JSONata for the conversion into logic program facts.
The endpoints of an express server ingested through a \gls{REST} interface are called by a Python script via MarkLogic to receive the facts from the knowledge graph.
This Python script invokes the clingo \gls{ASP} tool chain to compute the stable models. 
Finally, a graph shows a scatter-matrix according to different performance indicators of the computed models.
Each model can be visualized as a global logistics configuration in a Python Dash Plotly environment. 

The discussed concept of co-design development in Section~\ref{sect:intro} includes considering new variants of the industrial system for the product optimized towards some key performance indicators. 
Here, we mainly focus on the modeling part of the industrial system, i.e.\ the logistics facts and their requirements. 
The real industrial system is about the logistics of constructing an aircraft from multiple components. 
The goal of the task is to build a complete aircraft. 
Each component can be constructed from smaller components at some production site. 
These components need to be transported to the production sites where they can be further processed, possibly via warehouses. 
Several production sites exist in various countries and are reachable via various transport means. 
Not all production sites can construct all components, not all components can be transported with all transport means and not all production or warehouse sites can be reached by all transport means.

Because of the obligation of non-disclosure, we discuss here an adapted and simplified logistics system, which covers all relevant components that are necessary to compute configurations for the real industrial system:
Figure~\ref{fig:logistics-example} shows a fictional map.
The {production locations} \aPa, \bPa and \cPa are {located in} \Ac, \Bc, and \Cc, respectively.
The {warehouse location}s \aHa and \bHa are located in \Ac and \Bc, respectively. 
Existing {transport means} are \plane, \truck and \ship.
Possible transport means at each location and {transport routes} between each two locations are denoted by a line: A continuous yellow line denotes a \truck route, a dashed blue line denotes a \ship route and a dotted green line denotes a \plane route. 
The black circles at the corner of the squares of the production locations denote which of the four {part}s (\partAbbrv1, \partAbbrv2, \partAbbrv3 and \partAbbrv4) is {produceable at} which site.
The production plan that tells us the order in which the parts have to be assembled is as follows: \partAbbrv1 is the final product and needs \partAbbrv2 for its assembly. \partAbbrv3 is necessary for the production of \partAbbrv2 and in turn, \partAbbrv4 is necessary for the production of \partAbbrv3.

\subsection{From Knowledge Graph to Logic Program Facts} \label{sect:knowledgegraph}

The data for the global logistics scenario was modeled as an \gls{RDF} knowledge graph.
The implementation follows the technology stack presented in Section~\ref{sec:semanticIntegration},
The knowledge is scattered over multiple different sources is collected in manually created spreadsheets for the main bulk of the data but also captured as unstructured data in the form of documents.
Additional assumptions of the system were not captured in any written form but were acquired via interviews with subject-matter experts for the industrial system.
As a first step we modeled the ontology to capture all the the necessary knowledge.


In the next step we instantiated the concepts from the ontology with the help of collected data from the subject-matter experts.
In order to minimize the workload we decided that not every fact had to be added to the knowledge graph but rather be created with the help of reasoning and \gls{SWRL} rules whenever possible.

For example since the number of production locations in our industrial network can change, routes between these locations would have to be updated not only to the new location but also extended to all the already existing locations in our knowledge graph.
This would mean an effort of creating and maintaining $n(n-1)/2$ for the $n$ location nodes in the network (\cite{rodrigue2017geography}).
These efforts multiply if different transport means (like ships, trucks, airplanes) are taken into account, that all travel on different routes.

In order to better reflect the actual industrial system --- and to reduce the number of possible routes --- the knowledge graph was extended so that some locations support restricted transportation means.
For example locations without a harbor cannot handle ships.
Additionally, certain transportation means are only available in certain geographic regions of the world.
Further we implemented concepts for continental and intercontinental transport means to accurately model the \gls{SWRL} rules for overseas transport routes.
This had to be done due to the \glsdesc{OWA} used for reasoning.

Since the ontology has no means to accurately determine the distance between locations for each transport mean an attached routing service calculates them before uploading the knowledge graph into MarkLogic.

Finally, reasoning on the properties of the ontology was preformed to materialize additional facts.
Utilizing the \texttt{owl:inverseOf} object property automatically creates the reverse connection between entities.
The knowledge graph only indicates which transport means can carry certain parts (\texttt{can\_transport}).
The reasoner then materializes the inverse connection: which part can be transported by what transport mean (\texttt{is\_transported\_by}).
This was implemented in order to create shortcuts in the knowledge graph to easier extract the data for the creation of the logic program facts.

The goal was to provide as much information prior to the grounding of the logic program. This reduced the facts that had to be derived in \gls{ASP} and improved its performance for the model computation.
It also provides a mechanism to perform consistency checks while modeling with the \textit{Protégé} graphical editor.

\begin{figure}[t]
    \centering
    \includegraphics[width=\textwidth]{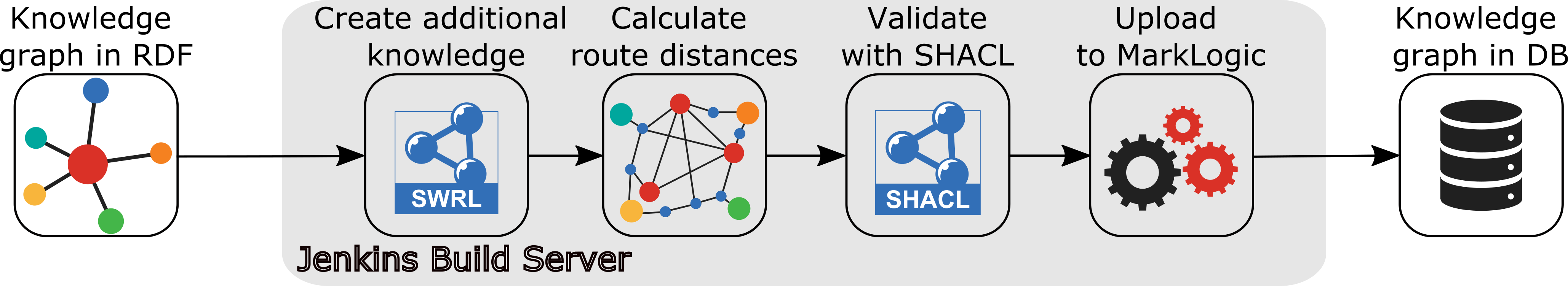}
    \caption{\gls{CI} process for deploying the global logistics knowledge base.
    \label{fig:kg-ic-process}}
\end{figure}

The data for the \gls{ASP} client was accessed through a MarkLogic data store (Section~\ref{sec:semanticIntegration}).
In order to reduce the amount of work the knowledge graph developer has to put into the deployment in MarkLogic a Jenkins task (\cite{cdf2023jenkins}) which is deployed to automatically push the data to the data store.
This happens every time a new push to the git repository is detected by a webhook (\cite{lindsay2007webhook}).
Before the data is ultimately pushed into the store several scripts are being executed by Jenkins.
First the different files are combined into a single file, for easier processing.
Then all the \gls{SWRL} rules are executed to materialize the additional facts.
Next, the data is completed by accessing the \gls{REST} service that calculated the distance between two locations.
As a last step a process checks via predefined \gls{SHACL} (\cite{w3c2017shacl}) shapes if all the entities in the knowledge graph conform to predefined shapes.
If all the entities conform to the \gls{SHACL} shapes they will be pushed into the data store.
Figure \ref{fig:kg-ic-process} gives an overview over the process.

To extract the data from the data store we use the provided \gls{SPARQL} endpoint of MarkLogic (Section~\ref{sec:semanticIntegration}).
For each set of logic program facts
a \gls{SPARQL} SELECT query is implemented.
\gls{SPARQL} SELECT queries produce a tabular output which needs to be transformed into such facts so they can be consumed by the clingo client.
For this we deploy a facade of \gls{REST} web services with Node.js that uses the JSONata library (\cite{jsonata2021jsonata}) to convert the resulting tabular data from the SELECT query into logic program facts.
Table~\ref{tab:lp-ontology-map} shows a mapping from the entities in the knowledge graph to the facts and gives one example for each (predicate) type.\footnote{In Figure~\ref{fig:logistics-example} \productionPart is abbreviated with \partAbbrv.}

\begin{table}
\begin{tabular}{@{\hspace{0cm}}l@{\hspace{0.2cm}}l@{\hspace{0.2cm}}l}
\toprule
\small
Entities in Knowledge Graph & Logic Program Fact Example & Description \\ \midrule
Instance of \texttt{Country} class & $\country(\Ac).$ & \text{\Ac is a country} 
\\\midrule
Instance of \texttt{ProductionLocation}   & $\prodLoc(\aPa).$ & \text{\aPa is a }\\
class and its sub-classes \texttt{Locations} && production location\\
 that can manufacture \texttt{Products} 
\\\midrule
Instance of \texttt{WarehouseLocation}  &
$\wareLoc(\bHa).$ & \text{\bHa is a}\\
classes and its sub-classes  && warehouse location
\\\midrule
Instance from \texttt{is\_located\_in} &		
$\locatedIn(\aPa, \Ac).$ & \text{\aPa is located}
\\ \texttt{Locations}
 in \texttt{Country}   &&  in  \Ac 
\\\midrule
Entities that can transport parts     &
$\transportMean(\ship).$ & \text{\ship is a}
\\
via \texttt{TransportationResource} &&  transport mean 
 \\\midrule
 via \texttt{has\_terminal} and \texttt{can\_handle} &
$\transportMeanAtSite(\bHa,\ship).$ & \text{\ship is a} \\
 what  \texttt{TransportationResource} &&  transport\\
 accepted at location&&  mean at \bHa
\\\midrule
Correlated to instances of \texttt{Product} &
$\productionPart(\productionPart2).$ & $\productionPart2$\text{ is a \productionPart} \\
 class and its sub-classes &&  to be produced
\\ \midrule
What \texttt{TransportationResource} &
$\canTransport(\ship,\productionPart2).$ & \text{\ship can } \\
can transport what \texttt{Product} && transport $\productionPart2$ \\\midrule
Via \texttt{ProductionResource} that links
		& $\partProduceableAt(\productionPart2,\aPa). $& $\productionPart2$\text{ is}
\\  via \texttt{has\_location} and \texttt{can\_produce}  &&  produceable at \aPa \\
\midrule
\texttt{Routes} via \texttt{has\_destination},
		&$\transportRoute(\aHa,\bHa,\ship,7).$ & \text{Distance from \aHa}  \\
 \texttt{has\_source} and \texttt{has\_transport\_mean} 
	& & to \bHa by \ship  is 7 
\\\midrule
combination of \texttt{Product} instances
& $\prodPlan(\productionPart1,\productionPart2).$ & $\productionPart2$ \text{necessary} for \\
 and \texttt{has\_part} object properties &&   production of $\productionPart1$ \\ \bottomrule
\end{tabular}
\caption{One example for each type of fact that is extracted from the knowledge graph. 
\label{tab:lp-ontology-map}
} 
\end{table}%

In the next section we describe how the facts from the industrial system knowledge graph are used for the definition of requirements in ASP.

\subsection{Representation of Logistics Requirements in ASP} \label{sect:lp}

\paragraph{Facts}
Eleven different (predicate) types of logic 
program facts with in total more than 70.000 facts, are extracted from the knowledge graph as described in Section~\ref{sect:knowledgegraph}. 


\paragraph{Rules}
The assembly of the aircraft is specified by a production plan (last row in Table~\ref{tab:lp-ontology-map}). 
For the example introduced at the beginning of this section,
$\productionPart1$ is 
the final product (or \rootPart). 
The following rule defines this \rootPart:
\begin{equation}
\rootPart(\Xvar) \leftarrow  \prodPlan(\Xvar,\_), \defneg \prodPlan(\_,\Xvar).
\end{equation}
For all $\Xvar$, if $\Xvar$ is a \prodLoc or \wareLoc, then $\Xvar$ is a \location:
\begin{equation}
 \location(\Xvar) \leftarrow \prodLoc(\Xvar). \quad\quad\quad
\location(\Xvar) \leftarrow \wareLoc(\Xvar).
\end{equation}
All \transportRoute relations are symmetric:
\begin{equation}
\transportRoute(\From,\To,\TM,\Distance) \leftarrow \transportRoute(\To,\From,\TM,\Distance). \label{r:transportroute}
\end{equation}
All locations have an \intrasiteTransport which can transport all parts 
with distance~$0$:
\begin{align}
&\transportMeanAtSite(\Xvar,\intrasiteTransport)\leftarrow  \location(\Xvar).\label{r:transportmeanatsite} \\
&\canTransport(\intrasiteTransport,\Xvar) \leftarrow \productionPart(\Xvar). 
\label{r:cantransport}\\
&\transportRoute(\Xvar,\Xvar,\intrasiteTransport,0) \leftarrow  \location(\Xvar). 
\label{r:transportrouteintra}
\end{align}
At all locations all parts can be transported intra-site with distance 0:
\begin{equation}
\small
\canBeTransportedFromTo (X,X,\variablePart,\intrasiteTransport,0) \leftarrow  \location(X), \productionPart(\variablePart). \label{r:canbetransportedfromto}
\end{equation}
A part can be transported from one location to another location by a certain transport mean with a given distance if the following holds:
\begin{align}
&\textsf{canBe}\textsf{TransportedFromTo} (\From,\To,\variablePart,\TM,\Distance) \leftarrow \canTransport(\TM,\variablePart),  \\
&\quad\quad
                                        \transportMeanAtSite(\From,\TM), \transportMeanAtSite(\To,\TM),
 \notag\\
&\quad\quad
\transportRoute(\From,\To,\TM,\Distance).  \notag
\end{align}
A \productionPart can be transported from its original \location (where it is produced) to its final \location (where it is further assembled), directly, via one or via two locations:
\begin{align}
&\direct (\variablePart,\From,\To,\TM,\Distance) \leftarrow  \canBeTransportedFromTo(\From,\To,\variablePart,\TM,\Distance). 
\medskip \\
& \via1 (\variablePart,\From,(\variableVia1,\To),(\TM1,\TM2),\Distance) \leftarrow 
\label{eq:via1}\\
& \quad\quad\canBeTransportedFromTo(\From,\variableVia1,\variablePart,\TM1,\Distance1),  \From != \To,  \notag
\\
&\quad\quad \canBeTransportedFromTo(\variableVia1,\To,\variablePart,\TM2,\Distance2), \Distance = \Distance1 + \Distance2.  \notag \medskip
\\
&\via2 (\variablePart,\From,((\variableVia1,\variableVia2),\To),(\TM1,\TM2,\TM3),\Distance)  \leftarrow  \label{eq:via2}
\\ 
&\quad\quad \canBeTransportedFromTo(\From,\variableVia1,\variablePart,\TM1,\Distance1),  \From != \To \notag
\\ 
&\quad\quad \canBeTransportedFromTo(\variableVia1,\variableVia2,\variablePart,\TM2,\Distance2), \variableVia1 != \variableVia2, \notag
\\
&\quad\quad  \canBeTransportedFromTo(\variableVia2,\To,\variablePart,\TM3,\Distance3), \Distance = \Distance1 + \Distance2 + \Distance3.\notag
\end{align}

%

\paragraph{Choice Rules} The requirements for the logistics system are specified through choice rules.
Each part needs to be produced at some production location at which it is produceable~\eqref{eq:single}
and each production location needs to produce at least one part~\eqref{eq:partProducedAtLocation}:
\begin{align}
1\{\partProducedAt(\variablePart,\PL):\partProduceableAt(\variablePart,\PL)\}1 \leftarrow \productionPart(\variablePart). \label{eq:single} \\
1\{\partProducedAt(\variablePart,\PL):\productionPart(\variablePart)\} \leftarrow  \prodLoc(\PL).  \label{eq:partProducedAtLocation}
\end{align}
These choice rules lead ASP to guess \partProducedAt(\variablePart,\PL) for each part and production location. 
After that, the transport paths and the transport means are guessed. 
If the part is the final product, i.e. the root part, then no further transportation is needed:
\begin{align}
&1\{\transportPath(\variablePart,\From,(),\intrasiteTransport,0):\partProducedAt(\variablePart,\From)\}1\leftarrow  \\
& \quad\quad \partProducedAt(\variablePart,\From), \rootPart(\variablePart). \notag
\end{align}
Otherwise, one of the other route types (\direct, \via1 or \via2) applies:
\begin{align}
&1\{\transportPath(\variablePart,\From,\To,\TM,\Distance):\direct(\variablePart,\From,\To,\TM,\Distance);\\
&\quad\quad
\transportPath(\variablePart,\From,(\variableVia1,\To),(\TM1,\TM2),\Distance): \notag\\
& \quad\quad\via1(\variablePart,\From,(\variableVia1,\To),(\TM1,\TM2),\Distance); \notag\\
& \quad\quad\transportPath(\variablePart,\From,((\variableVia1,\variableVia2),\To),(\TM1,\TM2,\TM3),\Distance):\notag\\
& \quad\quad\via2(\variablePart,\From,((\variableVia1,\variableVia2),\To),(\TM1,\TM2,\TM3),\Distance)\}1 :-\notag\\
&
                         \partProducedAt(\variablePart,\From), \productionPlan(\super,\variablePart), \partProducedAt(\super,\To). \notag
\end{align}

\paragraph{Integrity Constraint}

So far, we have assumed a single sourcing strategy, i.e.\ every part is produced at exactly one location. 
The resilience of a logistics system can be strengthened by applying a multi sourcing strategy. \label{multi-sourcing}
For double sourcing, every part is required to be produced at two different locations,
thus the min and max $1$ constraints in choice rule~\eqref{eq:single} need to be replaced by $2$. It is then desirable that these two locations are not located in the same country.
\begin{align}
&\leftarrow \productionPart(\variablePart),  \partProducedAt(\variablePart, \variableLocation1),
   \locatedIn( \variableLocation1, \variableCountry), \\
   & \partProducedAt(\variablePart, \variableLocation2),
    \locatedIn(\variableLocation2, \variableCountry),
    \variableLocation1 != \variableLocation2. \notag
\end{align}

\paragraph{Optimization}
The next statement minimizes the overall distance of the transport paths:
\begin{align}
& \#\minimize \{\Distance,\productionPart,\From,\To,\TM: \transportPath(\productionPart,\From,\To,\TM,\Distance)\}.
\end{align}
 
\paragraph{Assertions}
In total we defined 25 assertions, which are rules that test if the extracted facts are correctly specified.
Here are two examples: 
It should not be the case that a location is not located in a country~\eqref{eq:notsamecountry} and it should not be the case that locations are located in different countries~\eqref{eq:notdiffcountries}.
\begin{align}
& \invLocatedIn(\Xvar) \leftarrow \location(\Xvar), \defneg \locatedIn(\Xvar, \_).\label{eq:notsamecountry}
\\
& \invalidLocatedInTwoCountries(\Xvar) \leftarrow \location(\Xvar), \country(\Yvar1), \country(\Yvar2), \label{eq:notdiffcountries}
\\ 
& \quad\quad\locatedIn(\Xvar,\Yvar1), \locatedIn(\Xvar,\Yvar2), \Yvar1 != \Yvar2. \notag
\end{align}
These assertions could also be expressed as integrity constraints. 
However,  in case an integrity constraint is violated,  ASP would simply lead to UNSAT.
The expression as assertion leads the head of the respective `invalid' rule to be true in all models and allows us to immediately identify invalid specifications. 

\begin{figure}[t]
	\centering
	\includegraphics[width=\textwidth]{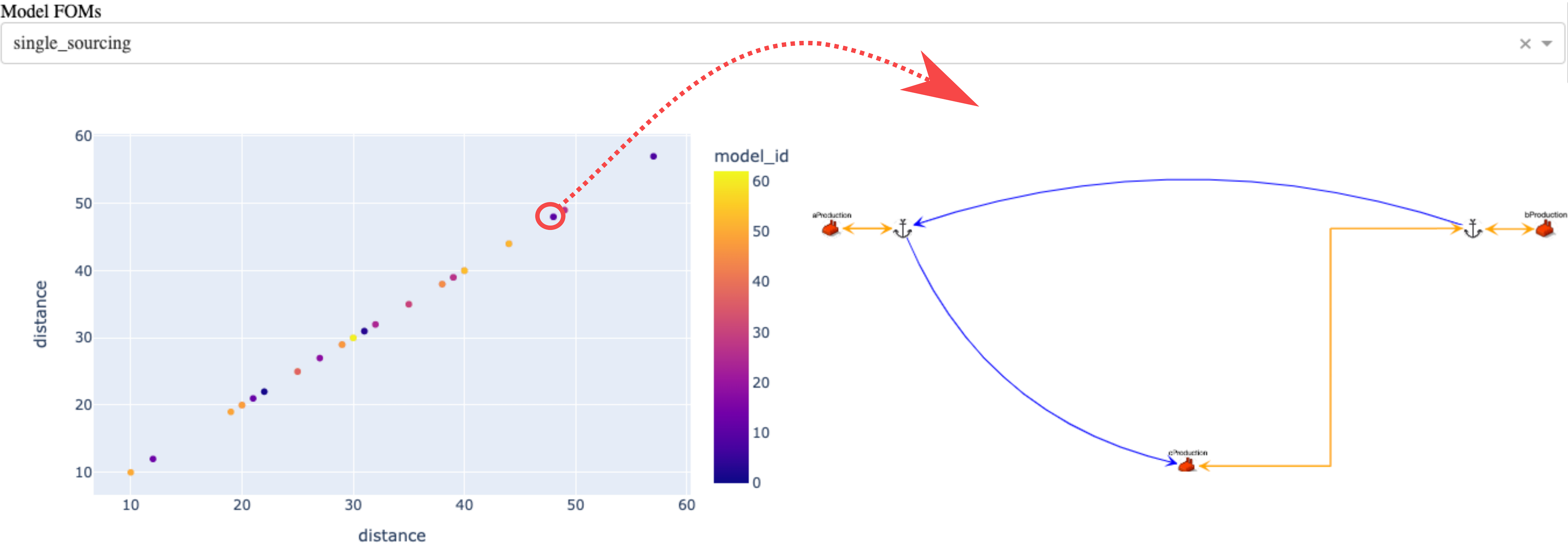}
	\caption{Overview of all models with respect to the total distance of necessary transportation.
	\label{fig:single-sourcing-overview}}
\end{figure}

\subsection{Visual Model Representation}\label{sect:visual}

A visual representation seems to be the simplest way for humans to communicate a model.
Post-processing was done with the computed models by translating the relevant facts into .csv tables and one overview table showing key performance indicators.
The graphical representation of the overview table and the individual configurations was done in a Python application using Plotly and Dash~(\cite{plotly,plotly-book}).
This toolkit produces an interactive representation of the already available .csv tables.
Figure~\ref{fig:single-sourcing-overview} shows an example for the simplified logistics system. The user interface is split into three parts:
The dropdown menu at the top shows that `single\_{sourcing}' is selected. 
When selected, a scatter matrix appears (left).
Each scatter plot represents the tradeoff between two indicators. 
As this example has only one performance indicator (`distance') it is simply plotted against itself. 
Each dot in the plot represents one computed model.
The user can then hover on any dot in order to show details of one model (right), which consists of a map, showing production locations (red small factories) and warehouses locations (black small anchors). 
The line between these locations represent instantiated routes, where colors and shapes of the lines denote the mode of transport. 
Hovering over each route shows the part that is being transported.
The red dotted arrow indicates that the red circled dot in the scatter matrix corresponds to the configuration shown in the map. This configuration shows a model with three indirect transport paths: \productionPart4 is transported from \cPa to \bPa via \bHa (anchor next to \bPa), \productionPart3 is transported from \bPa to \aPa via \bHa and \aHa and \productionPart2 is transported from \aPa to \cPa via \aHa (anchor next to \aPa) where \productionPart1 (root) is assembled into the final product.

\section{Variations \& Evaluation} \label{sect:lessonslearned}
Experiments on a 2.3 GHz Quad-Core Intel Core i7+16 GB RAM 3733 MHz type DDR4 with 29 locations (13 \prodLoc facts and 16 \prodLoc facts), 182 \transportMeanAtSite facts and 34 \productionPart facts (including one \rootPart), show that variations in the encoding can have a strong impact on grounding and solving time: 
Rules~\cref{r:transportroute,r:transportmeanatsite,r:cantransport,r:transportrouteintra,r:canbetransportedfromto} in Section~\ref{sect:lp} were replaced with almost 30.000 \canBeTransportedFromTo facts directly derived from the ontology, omitting the instantiation of irrelevant rules. The second row in Table~\ref{tab:eval},  Baseline,  shows the resulting number of (choice) rules, times (in minutes) for grounding and finding the first model.
The third row, PL Choice as IC, refers to the variation where choice rule~\eqref{eq:partProducedAtLocation} 
is replaced by the following rule and integrity constraint:
\begin{align}
& \partProducedAtLocation(\PL) \leftarrow \productionLocation(\PL), \partProducedAt(\variablePart, \PL). \\
&  \leftarrow \productionLocation(\PL),  \defneg \partProducedAtLocation(\PL).
\end{align}
The fourth row, Loc Type Req, refers to the extension of the \via1 \eqref{eq:via1} and \via2 \eqref{eq:via2} rules by requiring $\From, \To$ and $\variableVia1, \variableVia2$ to be production locations and warehouse locations, respectively. 
As all warehouses in the current ontology are harbours, the transport mean between \via1 and \via2 can only be a ship. 
The results by extending the \via2 rule to require $\TM2$ to be a ship,  i.e.\ $\ship(\TM2)$,  is shown in the fifth row, TM Type Req.
The last row,  All,  shows the results when all the replacements and extension were applied.

\begin{table}
\begin{tabular}{c c c c c} 
\toprule
LP Variation & \# Rules (\%) & \# Choice Rules (\%) & Grounding in min. & 1st model in min.\smallskip\\ \midrule
Baseline &  63148530 (100) & 4475  (100) & 14.47 & 5.67\\
PL Choice as IC & 63142310 (99) & 3098 (69) & 14.24 & 5.42\\
Loc Type Req & 2873012 (4.5) & 4307 (96) & 1.14 & 0.24\\
TM Type Req 
& 1313162 (2) & 3107 (69) & 0.19 & 0.05\\
All & 194129 (0.3) & 2667 (60) & 0.04 & 0.01\\
\bottomrule
\end{tabular}
\caption{\label{tab:eval}  Configurations wrt \# (choice) rules, time for grounding and 1st model.}
\end{table}

\section{Conclusions, Lessons Learned \& Future Work}\label{sect:conclusions}
The representation as \gls{RDF} ontology and the computation of configurations in \gls{ASP} seems promising for co-design development. 
With the help of the graphical Protégé editor it was easy to discuss the architecture, concepts and meta-data of the global industrial system with the industrial architects to crosscheck if we understood their provided constraints and data sets properly.
For the first time we were able to devise a machine readable model of the global industrial system.

The construction of additional facts with the help from \gls{SWRL} rules and reasoning was beneficial in two regards. 
Firstly, it allowed the creation of ancillary knowledge for the configuration of the logic programs.
It also provided a mechanism to quickly adapt the product and industrial system to new requirements from the stakeholders. 
Secondly, some entities 
like valid \texttt{Routes} for the global logistics system 
and additional facts
were generated automatically, which
reduced the number of entities in the graph that had to be added or adapted considerably.
Reasoning conducted under the \glsdesc{OWA} resulted in the creation of more concepts and properties to capture everything with the \gls{SWRL} and reasoning rules.
It was still manageable for our use case but could become a hindrance when scaling up the industrial system knowledge graph.
Alternative approaches like the \gls{OML}, that automatically create additional statements to close the world, could be deployed (\cite{maged2022oml}).
The \gls{CI} pipeline implemented with Jenkins made the deployment of new iterations of the knowledge graph seamless. 
Through the git webhook the \gls{CI} pipeline was always informed when new changes had to be deployed in MarkLogic.
The \gls{SHACL} shapes ensured that no invalid data was integrated into the knowledge graph.
The knowledge graph could locally implement changes while the computational intensive tasks
was handled by the Jenkins server.
The computed models in ASP were validated by a fully fledged simulation environment in AnyLogic~(\cite{tac2023anylogic}).


As Table~\ref{tab:eval} in Section~\ref{sect:lessonslearned} shows, adding \textit{obvious} information to the rules accelerated the search process in \gls{ASP} significantly:
The extension of the $\via1$~\eqref{eq:via1} and $\via2$~\eqref{eq:via2} rules with a requirement on the location type (row~4,  Loc Type Req) and a requirement on the transport mean type between warehouses (row~5, TM Type Req),  reduced grounding from minutes to seconds. 
A surprising feature of \gls{ASP} compared to other approaches that we investigated so far, is that it was very simple to specify partial configurations, when they were known (by simply adding them as facts) or modify requirements,  as shown by the simple extension for the multi sourcing strategy in Section~\ref{sect:loginaction} (on page~\pageref{multi-sourcing}).

In the future,  we intend to extract certain requirements directly from OWL,  such that choice rules and integrity constraints can be automatically built from templates.
Another aspect would be to produce a wider range of variations by means of the multishot feature in \gls{ASP}.
\cite{GRS2016} suggest the fixed-domain
semantics,  an alternative formal semantics for an intuitive understanding in OWL and description logics that might be accessible for logically less skilled practitioners and provide a translation into ASP. 
Even though our purpose was to reduce the grounding time by deriving relevant facts from the ontology directly,  their approach might be interesting for us. 

\section*{Acknowledgments}
Thanks to the anonymous reviewers for valuable feedback. It helped improving the paper.


\nocite{*}
\bibliographystyle{eptcs}
\bibliography{global-logistics}
\end{document}